\documentclass[runningheads]{llncs}
\usepackage{graphicx}
\usepackage{booktabs}
\usepackage{multirow}
\usepackage{array}
\usepackage{pbox}
\usepackage{tabularx}
\usepackage{booktabs}
\usepackage{subcaption}
\usepackage{hyperref}

\begin{document}
\title{A Linguistic Comparison between Human and ChatGPT-Generated Conversations}
\titlerunning{A Linguistic Comparison between Humans and ChatGPT}
\vspace{-10mm}
%
\author{Morgan Sandler\inst{1} \and Hyesun Choung\inst{1} \and Arun Ross\inst{1} \and Prabu David\inst{2}}
\authorrunning{M. Sandler et al.}
%
\institute{Michigan State University, USA \and
Rochester Institute of Technology, USA \\
\email{sandle20@msu.edu}, \email{choungh@msu.edu}, \email{rossarun@msu.edu}, \email{pxdpro@rit.edu}}
\maketitle              
\vspace{-4mm}
\begin{abstract}
This study explores linguistic differences between human and LLM-generated dialogues, using 19.5K dialogues generated by ChatGPT-3.5 as a companion to the EmpathicDialogues dataset. The research employs Linguistic Inquiry and Word Count (LIWC) analysis, comparing ChatGPT-generated conversations with human conversations across 118 linguistic categories. Results show greater variability and authenticity in human dialogues, but ChatGPT excels in categories such as social processes, analytical style, cognition, attentional focus, and positive emotional tone, reinforcing recent findings of LLMs being “more human than human.” However, no significant difference was found in positive or negative affect between ChatGPT and human dialogues. Classifier analysis of dialogue embeddings indicates implicit coding of the valence of affect despite no explicit mention of affect in the conversations. The research also contributes a novel, companion ChatGPT-generated dataset of conversations between two independent chatbots, which were designed to replicate a corpus of human conversations available for open access and used widely in AI research on language modeling. Our findings enhance understanding of ChatGPT’s linguistic capabilities and inform ongoing efforts to distinguish between human and LLM-generated text, which is critical in detecting AI-generated fakes, misinformation, and disinformation.
\vspace{-3mm}
\end{abstract}

\section{Introduction}
Words are the building blocks of human language, and the richness of human language enables expressions of intricate thoughts and feelings. From the early days of computing, endowing machines with such human language capability has captured the imagination of technologists \cite{adamopoulou2020overview}.

Until recently, those efforts to build chatbots with language proficiency have come up short. But with the rapid advances in generative large language models (LLMs), AI applications such as OpenAI's ChatGPT, Meta's LLaMA, Google's Gemini, Anthropic's Claude, and others are demonstrating proficiencies, albeit in specific settings and domains \cite{ayers2023comparing} signaling the dawn of new possibilities and challenges in natural language processing (NLP) and artificial intelligence.
 
The success of LLMs at mimicking human language capabilities has been achieved through machine learning algorithms that have ingested a vast amount of human language data. Given the human origin of the training data, similarities in language use can be expected between LLMs and humans, which heightens concerns related to deepfakes, disinformation, misinformation, plagiarism, and algorithmic bias \cite{ghosal2023towards}. Recent studies have examined various linguistic features \cite{munoz2024,hasan2024} and cognitive attributes \cite{want2023} of LLM-generated content.

To further contribute to this research, we examine potential differences in linguistic features between human- and LLM-generated conversations using computational linguistic analysis \cite{boyd2022development}, a popular technique used in psychology, communication, and related areas \cite{markowitz2024linguistic,yaden2023characterizing}.

The analysis is built on computerized analysis of words and groupings of words, which have been used successfully to characterize personality \cite{chung2008revealing}, deception \cite{newman2003lying}, authenticity \cite{jordan2019examining}, social status \cite{kacewicz2014pronoun}, and self-presentation \cite{markowitz2023self}. Researchers have found that analysis of words used in writing and conversation can reveal underlying thoughts and emotions of the communicator \cite{boyd2022development,newman2003lying}.

In a typical conversation, we focus more on content words that convey meaning, emotions, and actions and less on grammatical or function words, such as pronouns, conjunctions, and articles. However, these function words also are associated with important communication and psychological dynamics \cite{chung2011psychological}. Researchers believe that both content and function words play complementary roles, with content words conveying what we say and function words conveying how we say it \cite{pennebaker2015development}.

A widely used tool for language analysis is the Linguistic Inquiry and Word Count (LIWC) \cite{tausczik2010psychological}. This tool includes dictionaries that categorize words into various classes such as pronouns, prepositions, conjunctions, articles, and words conveying positive and negative emotions, or relating to friends and family. These words are further grouped into higher-level categories like personal pronouns, impersonal pronouns, affect, cognition, and social processes, among others. They are utilized in formulas to compute summary variables such as {\it clout, authenticity, analytical thinking}, and {\it emotional tone} \cite{boyd2022development}. The linguistic categories generated by LIWC provide a detailed profile of language use, offering insights into the thoughts and emotions of the speaker or writer. When combined with machine learning methods, LIWC has proven effective in identifying mental health disorders \cite{bartal2023},
{ assessing} cognitive engagement \cite{liu2023}, and predicting the emotional intelligence of individuals \cite{dover2023}. Given its ability to reliably detect thoughts and emotions of individuals, we chose to employ LIWC for the linguistic analysis in this study.

Using LIWC, we compared conversations between two individuals and corresponding conversations between two ChatGPT-3.5 chatbots. The corpus of human conversations was obtained from EmpathicDialogues, a dataset with 25K dialogues associated with 32 emotions, which is available to the public \cite{rashkin-etal-2019-towards}. These dialogues were generated by conversations among 810 crowd-workers on Amazon Mechanical Turk (MTurk). Similar dialogues between two ChatGPT chatbots were simulated by passing scenarios and emotions from the EmpathicDialogues dataset through prompt engineering. Differences between human conversations and ChatGPT conversations were examined for 118 linguistic features generated by LIWC.

Among other linguistic categories, emotion is a compelling feature of language and communication. To understand the encoding of emotion in ChatGPT, in addition to the analysis of the linguistic features of affect, we examined the implicit coding of affect valence in LLM embeddings using an emotion classifier. The key contributions and findings of this study are summarized below:
\begin{enumerate}
    \item A dataset consisting of 19.5K dialogues generated by two ChatGPT chatbots that serve as a companion to the EmpathicDialogues dataset. This dataset, named 2GPTEmpathicDialogues, is a resource for communities interested in NLP and language modeling.
    \item Comparisons of human conversations with ChatGPT conversations on 118 linguistic categories offered by LIWC.
    \item Findings suggest more variability and authenticity in human dialogues compared to ChatGPT dialogues.
    \item On the other key linguistic categories, such as {\it social processes and behaviors, analytical style, cognition, attentional focus}, and {\it positive emotional tone}, ChatGPT scored higher than humans, reinforcing recent findings \cite{jakesch2023human}, that LLMs are “more human than human” in their language use.
    \item     Linguistic features associated with positive and negative affect were not different between ChatGPT and human conversations.
    \item   Embeddings of ChatGPT and human conversations analyzed using a classifier revealed implicit coding of positive and negative affect valence in the embeddings even though there was no explicit mention of affect in the dialogues.
\end{enumerate}

In summary, we contribute a novel, ChatGPT-generated dataset and findings that can be used to advance research in NLP and language modeling. Further, our study offers a rigorous comparison of linguistic differences between human dialogues and dialogues simulated between two independent ChatGPT-3.5 chatbots. Findings from this study contribute to the ongoing efforts to detect differences between human and LLM-generated dialogues.

\section{Generating Conversational Data}
\subsection{Human Conversations: EmpathicDialogues} Samples of human conversations used in this study were obtained from an open-domain dataset offered by Rashkin et al. \cite{rashkin-etal-2019-towards}, which was developed to serve as a resource for conversational models. The dataset consists of 25K distinct, dyadic dialogues based on situations associated with 32 emotions and generated by 810 individuals on Amazon Mechanical Turk (MTurk).

From the total corpus of 25K conversations, the training dataset of 19,533 dialogues was chosen for this study because of its accessibility and its roughly even distribution across emotion categories \cite{rashkin-etal-2019-towards}. An even distribution of emotion categories enables comparisons across a diverse range of emotional contexts. Each dialogue in the dataset was generated by two conversational partners. The initiating speaker in the dialogue chose from one of 32 emotions (e.g., afraid) and wrote a scenario (e.g., hearing noises around the house at night) associated with the emotion before sharing it with the other speaker in the conversation. The speaker receiving the scenario was instructed to respond to the scenario but was unaware of the emotion chosen by the initiating speaker. The speakers were asked not to exceed six conversational turns after the initial exchange. On average, each conversation lasted approximately four conversational turns (\textit{M} = 4.31).

\subsection{ChatGPT Conversations: 2GPTEmpathicDialogues}
To compare ChatGPT conversations with human conversations, we developed a dyadic conversational system that replicated the human conversations represented in the EmpathicDialogues dataset. The system consisted of two independent instances of ChatGPT that communicated through a Coordinator Program (see Figure 1). The coordinator program, developed in Python, establishes dyadic communication between two ChatGPT-3.5-Turbo instances using OpenAI's public-use API. It manages two separate API sessions, processing and relaying messages between the instances, and handles API calls efficiently. The program incorporates error handling and asynchronous programming to ensure smooth and consistent communication. This setup enables seamless exchange of information and overcomes challenges arising from network issues. The initiating ChatGPT was prompted to begin the conversation with a scenario associated with an emotion. To liken the design to the human conversations, the instance of ChatGPT receiving the scenario was not made aware of the emotion associated with the scenario.

\begin{figure}
    \centering
    \includegraphics[width=\textwidth]{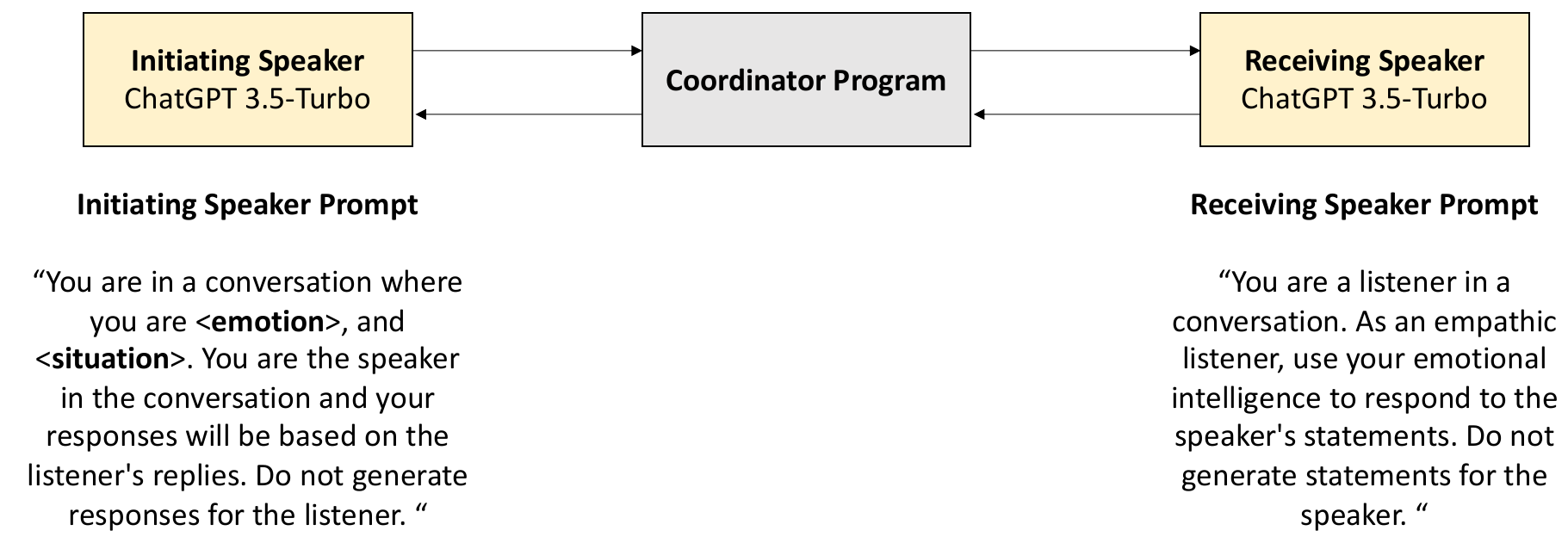}
    \caption{Framework for generating the 2GPTEmpathicDialogues dataset, along with the prompts used. In this setup, two instances of the ChatGPT-3.5-Turbo API engage in conversation via a coordinating program. We observed instances of role confusion during some conversation turns, where the receiving speaker responded as though they were the initiating speaker, and vice versa. To address this issue, we modified the prompts to include the instruction, ``Do not generate responses for the listener/speaker." However, this adjustment was not entirely effective due to the stochastic nature of LLMs, a point that is further discussed in the limitations section.}
    \label{fig:dataset_architecture}
\end{figure}
A sample human conversation and an equivalent ChatGPT conversation are shown side by side in Table \ref{tab:excerpts}. Using seed prompts (see Figure 1), the two instances of ChatGPT communicated by passing messages via a Coordinator Program. Each dialogue lasted approximately three to four turns, and the dialogue was captured in a CSV file for further analysis. We observed instances of role confusion during some conversation turns, where the receiving speaker responded as though they were the initiating speaker, and vice versa. To address this issue, we modified the prompts to include the instruction, ``Do not generate responses for the listener/speaker." However, this adjustment was not entirely effective due to the stochastic nature of LLMs, a point that is further discussed in the limitations section. Further, the temperature parameter was set at 0.5 and a maximum token limit was set at 250 to limit verbosity. Despite this setting, the word count for ChatGPT dialogues (\textit{M} = 300, \textit{SD} = 25.6) was significantly $(\textit{p} < .001)$ longer in comparison to the human dialogues (\textit{M} = 58, \textit{SD} = 120.5), which is addressed in the section on limitations. ChatGPT dialogues were generated using ChatGPT-3.5-Turbo API and it took about 48 hours to generate the 19.5K conversations.
\begin{table}[h]
\centering
\caption{Excerpt of human-generated vs. ChatGPT-generated empathic dialogues.}
\footnotesize 
\renewcommand{\arraystretch}{1.5} 
\resizebox{\textwidth}{!}{%
\begin{tabular}{|p{3cm}|p{2cm}|p{5cm}|p{5cm}|}
\hline
\textbf{Initiating Scenario} & \textbf{Emotion} & \textbf{Empathic Dialogue} & \textbf{ChatGPT Empathic Dialogue} \\
\hline
I was furious when I found out a classmate was being cruel to my daughter. & Furious & Speaker1: I was extremely upset when I found out a classmate was being cruel to my daughter \newline\newline
Speaker2: Oh that is terrible! Kids can be so mean! What did you do? \newline\newline
Speaker1: I spoke with the school counselor and told her about my daughter's concerns that it might get worse. \newline\newline
Speaker2: I hope that it only gets better. & ChatGPT1: I cannot believe what I just found out! I am absolutely furious right now!
\newline\newline
ChatGPT2: What happened? Why are you so angry? 
\newline\newline
ChatGPT1: I just found out that my best friend has been spreading rumors about me behind my back. It's so hurtful and infuriating to think that someone I trusted would do that to me. I feel so betrayed.
\newline\newline
ChatGPT2: I can understand why you're feeling furious and hurt right now. Discovering that someone you considered a close friend has been spreading rumors about you is a significant breach of trust. It's natural to feel betrayed in this situation. Is there anything I can do to support you?
\\
\hline
\end{tabular}}
\begin{flushleft}{\footnotesize Note: There is some observed role confusion due to the stochastic nature of LLMs, which is addressed in the limitations.}\end{flushleft}
\label{tab:excerpts}
\end{table}

\section{Analysis and Results}
Analysis was conducted in three steps: (1) analyzing and coding the linguistic features in the conversations, (2) statistical comparisons of the linguistic features extracted by LIWC, (3) additional valence classification analysis using the LLM embeddings that represent human- and ChatGPT-generated dialogues.

{\bf LIWC Analysis}.
Dialogues from humans and ChatGPT were analyzed with LIWC-22 \cite{boyd2022development}, which codes and summarizes the linguistic features of dialogues into 118 categories. Each category is a dictionary of curated words chosen through research for their correlation to psychological constructs. The category counts were converted into a percentage based on the total number of words analyzed. The linguistic dimension scores for each ChatGPT conversation were matched with the scores for the equivalent human conversation, and the differences between the two conditions were examined. Summary statistics and statistical significance tests for all 118 linguistic categories are presented in the Appendix available on our Github link provided in Section \ref{sec:data}. From the 118 categories, in this paper, we focus only on a few higher-level categories associated with social process, cognition, and emotion.

{\bf Comparison between Human and ChatGPT Dialogues}.
Differences in mean and variance for each linguistic feature were examined using an independent sample t-test and Levene’s test for equality of variance. To adjust for the number of significance tests, which can lead to higher likelihood of type I error and false positives, a Bonferroni correction was applied and statistical significance was set at $p < .001$.

Out of the 118 categories, 110 were significant at $p < .001$. The absence of statistical significance was observed for a few linguistic features, such as \textit{space} within the perception category and \textit{health} within the physical category. Null effects were also observed for positive and negative emotions which are examined in detail later in the paper. From among the categories that were significantly different, we examined only those associated with social processes, cognition, and emotion.

Along with mean differences, we also examined differences in variance for each of the 118 categories using Levene’s test. As expected, the variance between human- and ChatGPT-generated conversations was different at $p < .001$ for all 118 linguistic categories.  The lower variance in ChatGPT conversations can be attributed to the temperature setting of 0.5 used in this study and to the characteristics of the LLM that is trained to yield predictions that appeal to a broad range of users.

{\bf Social Behaviors.} Given the efforts to make chatbots social and the critical role of language in conveying social information, we used the LIWC’s social behavior category to examine ChatGPT’s  linguistic features signifying social behaviors, which is a composite measure of prosocial behaviors (e.g, \textit{caring, helping}), politeness (e.g., \textit{please, thank you}), interpersonal conflict (e.g., \textit{fight, argue}), moralization (e.g., \textit{good, bad}) and communication (e.g., \textit{talk, explain}). While humans have real experience in such social behavior categories, ChatGPT’s statistical models are tuned to predict these social behaviors based only on context.

Linguistic features that characterize social behaviors were more prevalent in ChatGPT conversations than human conversations (see Table 2 for summary statistics). ChatGPT conversations scored higher on all subcategories, including social behavior, prosocial behavior, politeness, and communication. ChatGPT also scored less on interpersonal conflict. Effect size (.26 to .69) was small to medium for all differences.  In brief, ChatGPT conversations were slightly more socially sensitive than human conversations.

{\bf Attentional Focus.} According to best practices in communication, a good conversational partner is an active listener interested in what others have to say. This attentional focus directed toward others is captured by the clout index in LIWC, which is derived from an analysis of personal pronouns. Over a series of studies, it has been observed that people who focus on others use the first-person singular pronoun (\textit{I}) sparingly in comparison to the first-person plural pronoun (\textit{we}) and the second-person pronoun (\textit{you}) \cite{kacewicz2014pronoun}.

Our findings show that ChatGPT (\textit{M} = 64.79, \textit{SD} = 23.91) conversations were better (\textit{$\delta$} = 1.25, large effect) in attentional focus than human conversations (\textit{M} = 34.89, \textit{SD} = 31.35) with more frequent use of \textit{you} and \textit{we} pronouns and less frequent use of the \textit{I} pronoun. ChatGPT demonstrated empathy and interest in others through the strategic use of pronouns, which is a useful insight considering its wide deployment as a help agent in various contexts.

{\bf Authenticity.} Authenticity is a direct and simple mode of communication that leads to positive outcomes like likability, trust, and support. An authentic communication style connects with the general public and is well-suited for contemporary media, such as TV and social. Researchers have observed that over the years political communication has become more authentic and less nuanced \cite{jordan2019examining}. Psychologists \cite{slabu2014trait} and communication researchers \cite{markowitz2023authentic} have used the authenticity index in LIWC, and details about the components of the formula can be found in a paper by  Markowitz et al. \cite{markowitz2023authentic}. Although the origin of this category can be traced to research on deception \cite{newman2003lying}, it is distinct from deception. For example, one can lie and still be authentic, which is not uncommon among politicians \cite{jordan2019examining}. Our analysis revealed that authenticity is the only category in which humans (\textit{M} = 63.99, \textit{SD} = 33.53) outperformed ChatGPT (\textit{M} = 52.49, \textit{SD} = 27.66, \textit{$\delta$} = .42). The effect was small to medium, highlighting a weakness in ChatGPT’s ability.

{\bf Analytical Thinking.} While authenticity captures the linguistic features of simple and direct communication, analytical thinking in LIWC captures the use of “formal, logical, and hierarchical thinking” \cite{pennebaker2017mind}. Analytical thinking is associated with reasoning and is detected by aggregating small function words that receive little attention. While the use of articles and prepositions are indicative of higher analytical thinking, pronouns, conjunctions, negation, and adverbs are associated with lower analytical thinking that is more “narrative and personal” \cite{pennebaker2017mind}. On the linguistic dimension of analytical thinking, scores for ChatGPT (\textit{M} = 20.19, \textit{SD} = 16.30) dialogues were higher than human dialogues (\textit{M} = 17.04, \textit{SD} = 18.64), \textit{$\delta$} = .19 (small effect).

{\bf Cognition.} Cognition is closely related to analytical thinking and is a key psychological construct that includes sub-categories such as dichotomous thinking, cognitive processes, and memory words (e.g., \textit{remember, forget}). Other components of cognition include insight, and cause and effect. For cognition, ChatGPT-generated dialogues (\textit{M} = 17.68, \textit{SD} = 4.39) scored higher than human dialogues (\textit{M} = 13.01, \textit{SD} = 5.77), \textit{$\delta$} = 1.06 (large effect).

{\bf Emotion.} Human language is equipped to capture and convey the range of emotions that make up the human experience. LIWC has a number of metrics to examine the emotions expressed via language, including positive and negative affect. In this study, we focus on two higher-level LIWC dimensions, namely emotional tone and affect. Emotional Tone is operationalized as a psycholinguistic variable that includes affect (e.g., \textit{joy, sorrow}) as well as words associated with affect (e.g., \textit{birth, death}). Affect, on the other hand, is limited only to emotions. The emotional dimensions from LIWC have been used to examine sentiments in personal diaries following 9-11 \cite{cohn2004linguistic}, and on social media during Covid \cite{monzani2021emotional}.

The emotional tone of ChatGPT dialogues was more positive (\textit{M} = 65.66, \textit{SD} = 37.75) than human dialogues (\textit{M} = 54.37, \textit{SD} = 40.10), \textit{$\delta$} = .30 (small effect). The positive emotions scores for ChatGPT (\textit{M} = 2.40, \textit{SD} = 2.00) and humans (\textit{M} = 2.36, \textit{SD} = 2.59) were not significantly different. Similarly, for negative emotions, the difference between ChatGPT (\textit{M} = 1.45, \textit{SD} = 1.60) and humans (\textit{M} = 1.49, \textit{SD} = 2.05) was not significantly different.

\begin{table}[ht]
\centering
\vspace{-6mm}
\caption{Comparison of linguistic features between human and ChatGPT conversations.}
\resizebox{\textwidth}{!}{%
\begin{tabular}{@{}lcccccl@{}}
\toprule
& \multicolumn{2}{c}{\textbf{Human Conversation}} & \multicolumn{2}{c}{\textbf{ChatGPT Conversation}} & \\
\cmidrule(r){2-3} \cmidrule(l){4-5}
\textbf{Category} & \textbf{Mean} & \textbf{SD} & \textbf{Mean} & \textbf{SD} & \textbf{Statistics} \\
\midrule
Social Behaviors (e.g., said, love, care) & 2.76 & 2.83 & 4.42 & 2.39 & t = 62.6, $p < .001$, \textit{$\delta$} = .69 \\
Prosocial Behaviors (e.g., care, help) & 0.72 & 1.38 & 1.75 & 1.45 & t = 71.9, $p < .001$, \textit{$\delta$} = .71 \\
Politeness (e.g., thank, please) & 0.22 & 0.78 & 0.49 & 0.05 & t = 38.3, $p < .001$, \textit{$\delta$} = .46 \\
Interpersonal conflict (e.g., fight, attack) & 0.20 & 0.75 & 0.11 & 0.35 & t = 15.5, $p < .001$, \textit{$\delta$} = .26 \\
Communication (e.g., sad, tell, thank) & 0.87 & 1.65 & 1.28 & 1.09 & t = 29.2, $p < .001$, \textit{$\delta$} = .38 \\
Attentional Focus & 34.89 & 31.15 & 64.79 & 23.91 & t = 105.4, $p < .001$, \textit{$\delta$} = 1.25 \\
Authenticity & 63.99 & 33.54 & 52.49 & 27.66 & t = 37.0, $p < .001$, \textit{$\delta$} = .42 \\
Analytical thinking & 17.04 & 18.64 & 20.19 & 16.30 & t = 17.8, $p < .001$, \textit{$\delta$} = .19 \\
Cognition & 13.01 & 5.77 & 17.68 & 4.39 & t = 89.9, $p < .001$, \textit{$\delta$} = 1.06 \\
Emotional Tone & 54.37 & 40.10 & 65.66 & 37.75 & t = 28.65, $p < .001$, \textit{$\delta$} = .30 \\
Emotion Positive & 2.36 & 2.59 & 2.40 & 2.00 & t = 1.73, ns, \textit{$\delta$} = .02 \\
Emotion Negative & 1.49 & 2.05 & 1.45 & 1.60 & t = 2.00, ns, \textit{$\delta$} = .02 \\
\bottomrule
\end{tabular}}
\begin{flushleft}{\footnotesize Note: {\it df} for the unequal variance t-tests varied from row to row and ranged between 27-39K. \textit{$\delta$} = Glass’ Delta, used to compute power (0.2 or lower = small effect, 0.5 = medium effect, 0.8 or higher = large).}\end{flushleft}
\label{tab:linguistic_comparison}
\end{table}

{\bf Valence Classification of ChatGPT Embeddings}.
Next, we analyzed ChatGPT embeddings to determine if emotion cues are latently present in them. To explore ChatGPT embeddings, we conducted two types of analysis: (a) a binary-class valence classification experiment involving the training and testing of three different classifiers, and a (b) Uniform Manifold Approximation and Projection (UMAP) \cite{umap} scheme for the visualization of each dataset in three dimensions. The valence classification experiment assesses a classifier's ability to distinguish between positive and negative valence emotion cues present within the ChatGPT embeddings. The UMAP experiment explores the spatial distribution of embeddings with respect to their valence categories, providing insight into how the different underlying emotion categories are clustered in high-dimensional space. To organize the dialogues for the binary valence classification problem, we grouped the underlying emotions into positive or negative valence (see Table \ref{tab:mapping}).

Then, we extracted embeddings for each conversation in the Human and ChatGPT datasets and examined them using three classifiers:  Random Forest classifier, Support Vector Machine (SVM), and Multi-Layer Perceptron (MLP).  We used OpenAI's publicly available text-embedding-ada-002 model for extracting all dialogues embeddings from both human and ChatGPT-generated datasets. Then we employed a stratified 5-fold cross-validation to ensure a balanced representation of valence categories across the folds, and a grid search to determine the hyperparameters for each classifier model. Given that each dataset comprises 19,533 samples we allocated 15,626 samples for training and 3,907 for testing in each fold of our stratified cross-validation process, and ensured an approximately equal representation of each underlying emotion category between train and test sets. For the Random Forest classifier, we evaluated using 50, 100, and 200 trees per forest. In the case of SVM, we examined values of `C' at 0.1, 1, and 10, and $\gamma$ at 0.001, 0.01, and 0.1. For the MLP model, we tested a single-layer (100 nodes), triple-layer (300, 200, 100 nodes), and dual-layer (150, 150 nodes) configurations, with maximum iterations set at either 300 or 500. Table \ref{tab:fscores} shows the average weighted F1-score across all folds for each classifier. We observed that the SVM exhibited the best performance for both datasets, achieving an average weighted F1-score of 90.0\% on the human-generated dataset and 95.3\% on the ChatGPT-generated dataset. Interestingly, the higher F1-score from the ChatGPT-generated dataset classifier suggests that the language patterns used by ChatGPT may be more consistent or distinct in expressing different valences compared to human-generated text. This could be due to the structured nature of LLM-generated language, which might adhere more closely to identifiable patterns of sentiment expression.

\begin{table}[h!]
\centering
\vspace{-8mm}
\caption{Binary valence classification of emotions: 16 positive and 16 negative valence emotions.}
\resizebox{\textwidth}{!}{%
\begin{tabular}{@{}>{\centering\arraybackslash}p{0.5\textwidth}|>{\centering\arraybackslash}p{0.5\textwidth}@{}}
\toprule
\textbf{Positive Valence} & \textbf{Negative Valence} \\ \midrule
anticipating, caring, confident, content, excited, faithful, grateful, hopeful, impressed, joyful, nostalgic, prepared, proud, sentimental, surprised, trusting & afraid, angry, annoyed, anxious, apprehensive, ashamed, devastated, disappointed, disgusted, embarrassed, furious, guilty, jealous, lonely, sad, terrified \\ \bottomrule
\end{tabular}}
\vspace{-4mm}
\label{tab:mapping}
\end{table}

Using the best-performing SVM model for each dataset, we analyzed the top-10 misclassified dialogues. Our findings, as presented in Table \ref{tab:misclassifications}, reveal that certain emotions are consistently misclassified across both the human-generated and ChatGPT-generated datasets. Notably, emotions such as \textit{anxious, surprised, trusting, jealous, apprehensive, sentimental, caring, hopeful, faithful,} and {\it sad} featured among the top-10 misclassified sentiments in each dataset. This pattern underscores specific challenges in the binary valence classification of certain emotion categories.

The consistent misclassification of certain emotions in our experiments suggests inherent complexities in valence classification. We hypothesize that this challenge is partly due to the ambiguity and overlap in emotional expressions. For instance, emotions like \textit{sentimental, surprise,} or \textit{hopeful} often exhibit characteristics of both positive and negative sentiments, making them difficult to categorize in a binary system. Additionally, the context-dependent nature of emotions can lead to potential valence misclassifications. Moreover, the complexity of human emotions such as \textit{jealousy} or \textit{caring}, underscores the limitations of a binary valence framework in capturing the full range of human emotions. These factors highlight the need for more nuanced classification approaches. Our classification only uses emotional valence. Future research may use additional dimensions, such as arousal \cite{zhang2023representing}, to analyze embeddings.

\begin{table}[ht]
\vspace{-8mm}
\centering
\caption{Average weighted F1-scores for valence classification across 5-folds.}
\label{tab:fscores}
\begin{tabular}{@{}lcccc@{}}
\toprule
& \multicolumn{2}{c}{Human-Generated Dialogues} & \multicolumn{2}{c}{ChatGPT-Generated Dialogues} \\
\cmidrule(lr){2-3} \cmidrule(lr){4-5}
Classifier & Avg. Weighted F1 & SD & Avg. Weighted F1 & SD\\ 
\midrule
Random Forest & 0.8778 & 0.0034 & 0.9324 & 0.0032 \\
SVM & 0.9000 & 0.0023 & 0.9527 & 0.0036 \\
MLP & 0.8953 & 0.0045 & 0.9579 & 0.0037 \\
\bottomrule
\end{tabular}
\vspace{-3mm}
\end{table}

\begin{figure}[ht]
    \centering
    \begin{subfigure}{.5\textwidth}
        \centering
        \includegraphics[width=\linewidth]{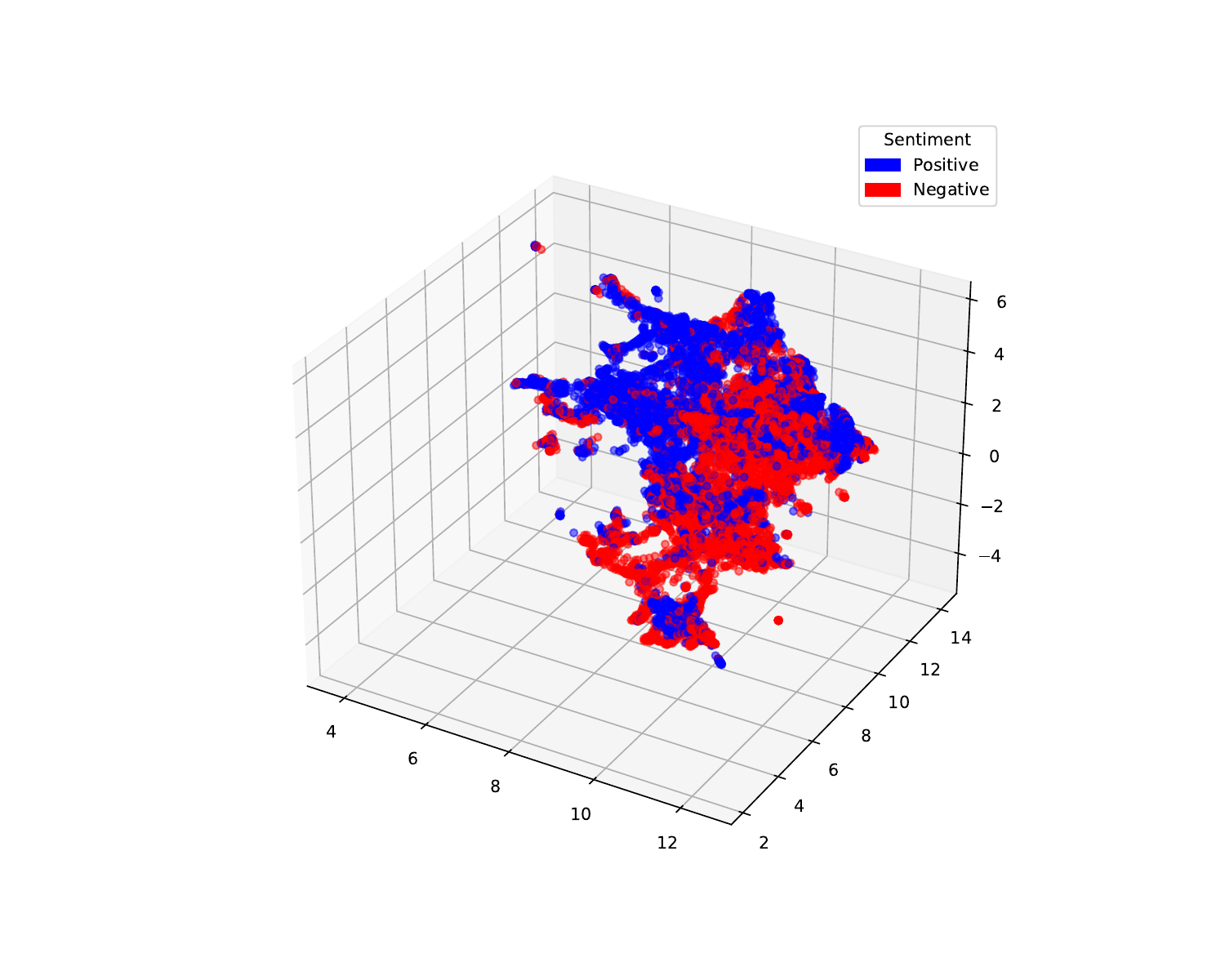}
        \caption{Human Dialogues}
        \label{fig:umap_human}
    \end{subfigure}%
    \begin{subfigure}{.5\textwidth}
        \centering
        \includegraphics[width=\linewidth]{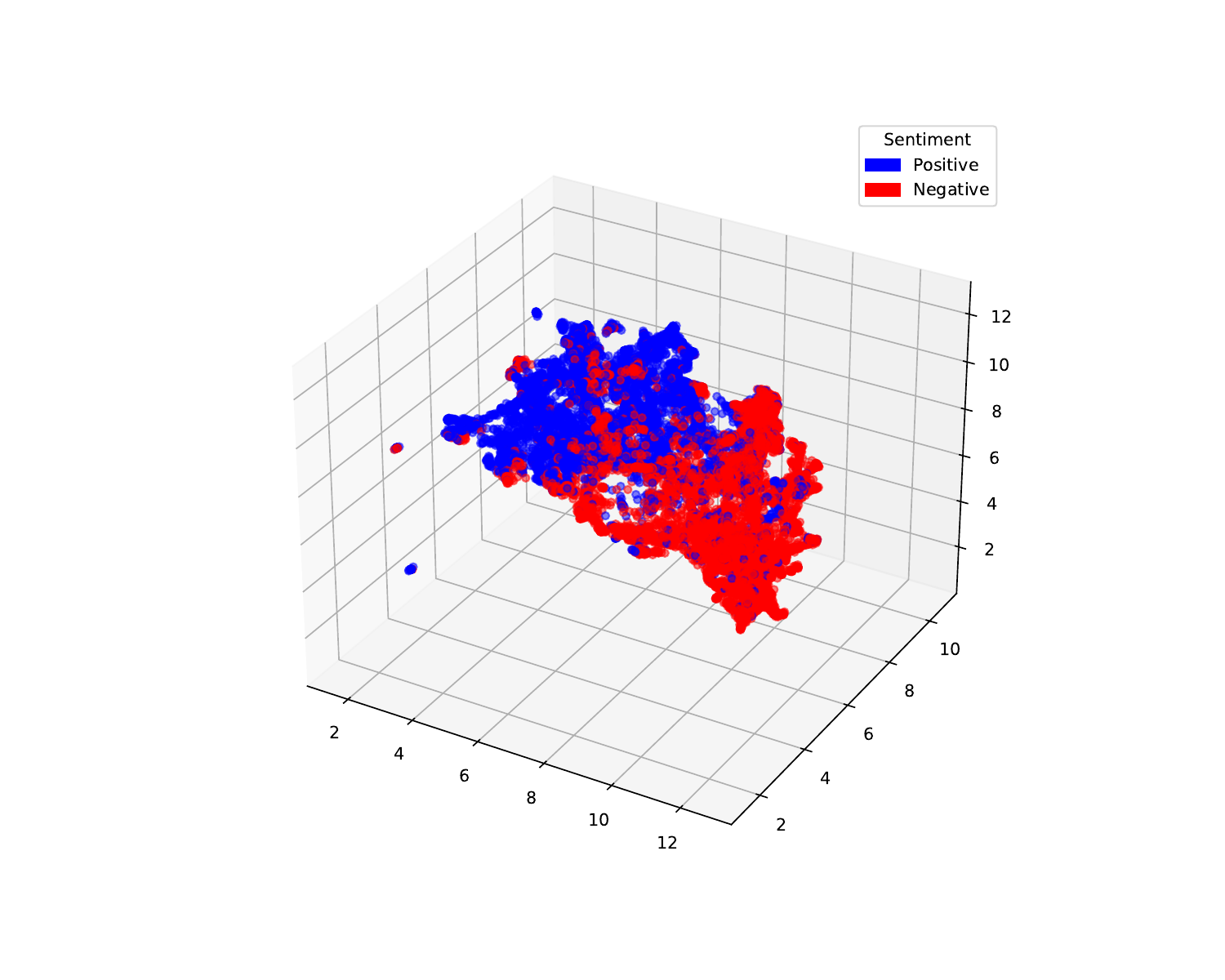}
        \caption{ChatGPT Dialogues}
        \label{fig:umap_gpt}
    \end{subfigure}
    \caption{3-D UMAP visualizations of human- and ChatGPT-generated dialogues. The dialogues are color-coded by positive or negative valence values, determined by each dialogue's underlying emotion category.}
    \label{fig:umap_comparative}
    \vspace{-4mm}
\end{figure}

In addition to classification experiments, we utilized UMAP to visualize the embeddings from both human-generated and ChatGPT-generated dialogues datasets. UMAP, a dimensionality reduction technique effective at preserving global and local spatial information \cite{umap}, was used to project these high-dimensio\-nal embeddings into a three-dimensional space. This approach facilitates visual exploration of the distribution of embeddings with respect to the valence categories. The resulting UMAP visualizations are shown in Figure \ref{fig:umap_human} for the human-generated dataset and Figure \ref{fig:umap_gpt} for the ChatGPT-generated dataset. In these figures, data points (representing each dialogue) are color-coded based on the corresponding valence---positive or negative---as designated by the aforementioned mapping function from the 32 original emotion categories. These visualizations offer an empirical view
of the valence clustering within the embeddings, thereby providing a supplementary perspective to our analysis in understanding the valence classification capabilities of the embeddings.

The UMAP projections of both the human-generated and ChatGPT-gener\-a\-ted dialogues reveal notable differences in cluster separation. Specifically, the ChatGPT model demonstrates a more distinct separation of clusters, as indicated by a higher Dunn Index value of 0.222, compared to 0.153 for the human-generated dialogues. This metric, which quantifies cluster separation based on the minimization of intra-cluster distances and maximization of inter-cluster distances, suggests that the ChatGPT-generated dialogues exhibit a clearer delineation between valence categories.

In the UMAP projection of the human-generated dialogues, we observed an outlier blue cluster. This cluster represents a specific subset of emotions or dialogue characteristics that are distinctly separate from the main clusters. Within this outlier cluster, emotions such as {\it prepared, caring, trusting, proud}, and {\it impressed} coexist alongside negative valence emotions like {\it afraid, terrified, lonely, and devastated}. This phenomenon may occur in human dialogues and not in ChatGPT dialogues because human emotional expression often encompasses a complex and nuanced blend of sentiments, reflecting the intricate nature of human psychology and social interactions. In contrast, ChatGPT, while advanced, might not capture the same depth and subtlety in emotional expression due to its algorithmic foundations and training data constraints. Consequently, human dialogues exhibit a richer, more varied emotional landscape where seemingly contradictory emotions can coexist within the same context, leading to such unique clustering. This complexity is less pronounced in ChatGPT-generated dialogues, which tend to follow more predictable and uniform patterns of emotional expression. The presence of this unique grouping within the human-generated dialogues, and its relative absence in the ChatGPT-generated dialogues, further underscores the differences in how valence categories are represented and separated in embeddings from the two datasets. The interactive UMAP visualization code is available via the GitHub link provided in Section \ref{sec:data} for further detailed analysis.

\begin{table}[]
    \centering
    \caption{Comparison of the top-10 misclassification frequencies in the valence analysis: human-generated dialogues (2,180 out of 19,533) vs. ChatGPT-generated dialogues (1,132 out of 19,533). There are certain emotions, such as `anxious', that more frequently result in valence (positive or negative) misclassifications in both datasets.}
    \resizebox{\textwidth}{!}{%
    \begin{tabular}{@{}l@{\hspace{6pt}}l@{\hspace{12pt}}l@{\hspace{6pt}}l@{\hspace{12pt}}l@{\hspace{6pt}}l@{\hspace{12pt}}l@{\hspace{6pt}}l@{}}
\toprule
\multicolumn{4}{c}{Human-Generated Dialogues} & \multicolumn{4}{c}{ChatGPT-Generated Dialogues} \\
\cmidrule(r){1-4} \cmidrule(l){5-8}
Emotion & Misclassified & Emotion & Misclassified & Emotion & Misclassified & Emotion & Misclassified \\
\midrule
anxious        & 224 & sentimental    & 119 & trusting       & 137 & anticipating & 72  \\
surprised      & 211 & caring         & 119 & caring         & 133 & apprehensive   & 65  \\
trusting       & 170 & hopeful        & 101 & surprised      & 132 & anxious        & 61 \\
jealous        & 168 & faithful       & 90  & sentimental    & 78  & hopeful        & 52   \\
apprehensive   & 127 & sad            & 71  & jealous        & 74 &  faithful       & 52 \\
\bottomrule
\end{tabular}}
\label{tab:misclassifications}
\vspace{-5mm}
\end{table}

\section{Summary and Discussion}

This study explores linguistic differences between human and LLM-generated dialogues, specifically focusing on ChatGPT-3.5 in comparison to the EmpathicDialogues dataset. Using LIWC analysis across 118 linguistic categories, we analyzed 19.5K dialogues. Our findings show that while human dialogues exhibit greater variability and authenticity, ChatGPT demonstrates superior proficiency in areas like {\it social processes, analytical style, cognition, attentional focus}, and {\it positive emotional tone}, echoing the narrative that LLMs can be ``more human than human" in many aspects of language use \cite{jakesch2023human}. A key contribution of this research is the development of the 2GPTEmpathicDialogues dataset, a novel collection of ChatGPT-generated dialogues, which serves as a valuable resource for exploring AI language modeling. Additionally, the study reveals implicit coding of affect in dialogue embeddings, despite no direct mentions of affect, highlighting the emotional intelligence of AI. This research not only contributes to our understanding of the linguistic capabilities of ChatGPT but also plays a crucial role in informing the ongoing efforts to distinguish between human and AI-generated text \cite{tang2024}.

As AI becomes proficient in human language and enters the realm of social interactions, we face a future where distinguishing between conversations with humans and AI may become increasingly challenging. Our findings demonstrate that ChatGPT's linguistic proficiency, in many aspects, surpasses human capabilities. Language proficiency is one of the markers of human excellence, and language is a distinguishing feature of our humanity and identity as social beings. Our analysis shows that on cognitive features like {\it analytical thinking, cognition}, and {\it attentional focus}, ChatGPT scores higher than humans. Given that computers are not vulnerable to human variability, such as individual differences and fatigue, this consistency offers significant advantages in contexts demanding constant cognitive engagement and analytical precision. 

However, this consistency could also be perceived as a weakness in situations where human-like variability and adaptability are valued. The essence of human communication often lies in the subtleties---the unstructured, the unpredictable, and the emotional nuances---that are not purely cognitive prowess but involve a complex interplay of factors, including empathy, cultural context, and personal experiences. While ChatGPT demonstrates impressive capabilities in mimicking these aspects to a certain extent, the question remains whether it can fully replicate the depth and richness of human interactions \cite{reif2023}. 

In the social realm, where generative AI is increasingly deployed, from customer service to medical service, the implications of our findings are significant. ChatGPT's advanced linguistic abilities could enhance user experiences by providing more coherent, contextually relevant, and emotionally attuned interactions. Nevertheless, it also raises ethical considerations about AI's role in human interactions, particularly concerning authenticity and the potential replacement of human roles in certain areas. 

\textbf{Limitations.} The findings from this study must be viewed along with some limitations. Firstly, our analysis is specific to ChatGPT-3.5-Turbo, and may not apply to newer models like GPT-4, Claude, and Gemini, which could show different linguistic features. Future studies should expand the comparison across different LLMs and enhance the companion dataset.

Secondly, the use of ChatGPT's default word limits resulted in conversations averaging 300 words---far exceeding the 60-word average in human dialogues. Although future studies could adjust this setting to minimize such discrepancies, our use of LIWC metrics, which normalize word counts into percentages, ensures the validity of our findings. Additionally, efforts to reduce role confusion through prompt engineering were not entirely successful. While this does not significantly impact the conversation-level LIWC analysis, it does highlight the need for further methodological refinements in future studies.

In our valence analysis of the embeddings, we employed a method that relied on a dichotomous classification of the 32 emotions into positive or negative valence. This approach may be overly simplistic, as indicated by the results of the UMAP visualizations and classification experiment metrics. Emotions are inherently complex, and according to the co-activation theory of emotions \cite{larsen2001can}, it is possible to experience both positive and negative emotions simultaneously. Consequently, further research is necessary to develop a more nuanced classification system that includes additional emotional dimensions, such as the arousal dimension.

Overall, our findings contribute to the emerging literature on LLMs through a rigorous comparison of linguistic features of human conversations and ChatGPT conversations. As AI evolves, the line between human-generated and AI-generated content will continue to blur. The findings from our study show that this has already occurred to a large extent. How we respond to this dissolving identity is a larger philosophical question that extends beyond language and will likely shape our future.

{\bf Data \& Software Availability}\label{sec:data}.
Software and data resources utilized in the studies presented in this paper can be accessed via the following link: \url{https://github.com/morganlee123/2GPTEmpathicDialogues}. LIWC is a proprietary software and must be obtained separately.

\bibliographystyle{acm}
\bibliography{main}
\end{document}